\documentclass[10pt,twocolumn,letterpaper]{article}

\usepackage[pagenumbers]{wacv} %

\usepackage{graphicx}
\usepackage{amsmath}
\usepackage{amssymb}
\usepackage{booktabs}
\usepackage{multirow,bm,caption,tabularx,colortbl,array,adjustbox,url}
\usepackage[dvipsnames]{xcolor}

\usepackage[pagebackref,breaklinks,colorlinks]{hyperref}

\usepackage[capitalize]{cleveref}
\crefname{section}{Sec.}{Secs.}
\Crefname{section}{Section}{Sections}
\Crefname{table}{Table}{Tables}
\crefname{table}{Tab.}{Tabs.}

\def\wacvPaperID{21} %
\def\confName{WACV}
\def\confYear{2025}

\newcolumntype{R}[2]{%
    >{\adjustbox{angle=#1,lap=\width-(#2)}\bgroup}%
    l%
    <{\egroup}%
}
\newcommand*\rot{\multicolumn{1}{R{45}{1em}}}

\definecolor{citecolor}{HTML}{0071BC}
\definecolor{linkcolor}{HTML}{ED1C24}
\definecolor{Gray}{gray}{0.86}

\begin{document}

\title{FLAVARS: A Multimodal Foundational Language and Vision\\Alignment Model for Remote Sensing}

\author{%
Isaac Corley$^1$\thanks{Corresponding author: \texttt{isaac.corley@utsa.edu}. Work done during internship with the Microsoft AI for Good Research Lab} \quad Simone Fobi Nsutezo$^2$ \quad Anthony Ortiz$^2$ \quad Caleb Robinson$^2$\\Rahul Dodhia$^2$ \quad Juan M. Lavista Ferres$^2$ \quad Peyman Najafirad$^1$\vspace{2mm}\\
$^1$University of Texas at San Antonio \quad \quad $^2$Microsoft AI for Good Research Lab
}

\maketitle

\begin{abstract}
Remote sensing imagery is dense with objects and contextual visual information. There is a recent trend to combine paired satellite images and text captions for pretraining performant encoders for downstream tasks. However, while contrastive  image-text methods like CLIP enable vision-language alignment and zero-shot classification ability, vision-only downstream performance tends to degrade compared to image-only pretraining such as MAE. In this paper, we propose FLAVARS, a pretraining method that combines the best of both contrastive learning and masked modeling, along with geospatial alignment via contrastive location encoding. We find that FLAVARS significantly outperforms a baseline of SkyCLIP for vision-only tasks such as KNN classification and semantic segmentation, +6\% mIOU on SpaceNet1, while retaining the ability to perform zero-shot classification, unlike MAE pretrained methods.
\end{abstract}

\section{Introduction}
\label{sec:intro}
Multimodal learning in remote sensing with paired images and text descriptions has enabled important applications such as text-to-image retrieval, image captioning, visual-question answering, and zero-shot classification. Using language as a method of understanding satellite and aerial imagery is a clever workaround to the potentially infinite number of unique objects on the surface of the Earth. The success of pretraining methods utilizing these datasets suggests that pretraining with multimodal data can yield relevant frozen representations and weight initializations for downstream tasks. However, while multimodal pretraining provides benefits over vision-only methods, e.g. zero-shot classification, this approach generally results in a trade-off which degrades performance on visually dense tasks such as segmentation or detection. El Banani et al.~\cite{el2024probing} studied the visual performance of a variety of foundation models and discovered that multimodal vision-language pretraining, such as CLIP~\cite{radford2021learning} and SigLIP~\cite{zhai2023sigmoid}, performs significantly worse for vision-only tasks such as depth estimation and geometric correspondence estimation. Singh et al.~\cite{singh2022flava} similarly proposed FLAVA which improved this trade-off by combining masked-image-modeling (MIM), masked-language-modeling (MLM), and contrastive learning in a unified pretraining framework.

\begin{figure*}[t!]
    \centering
    \includegraphics[width=1.0\linewidth]{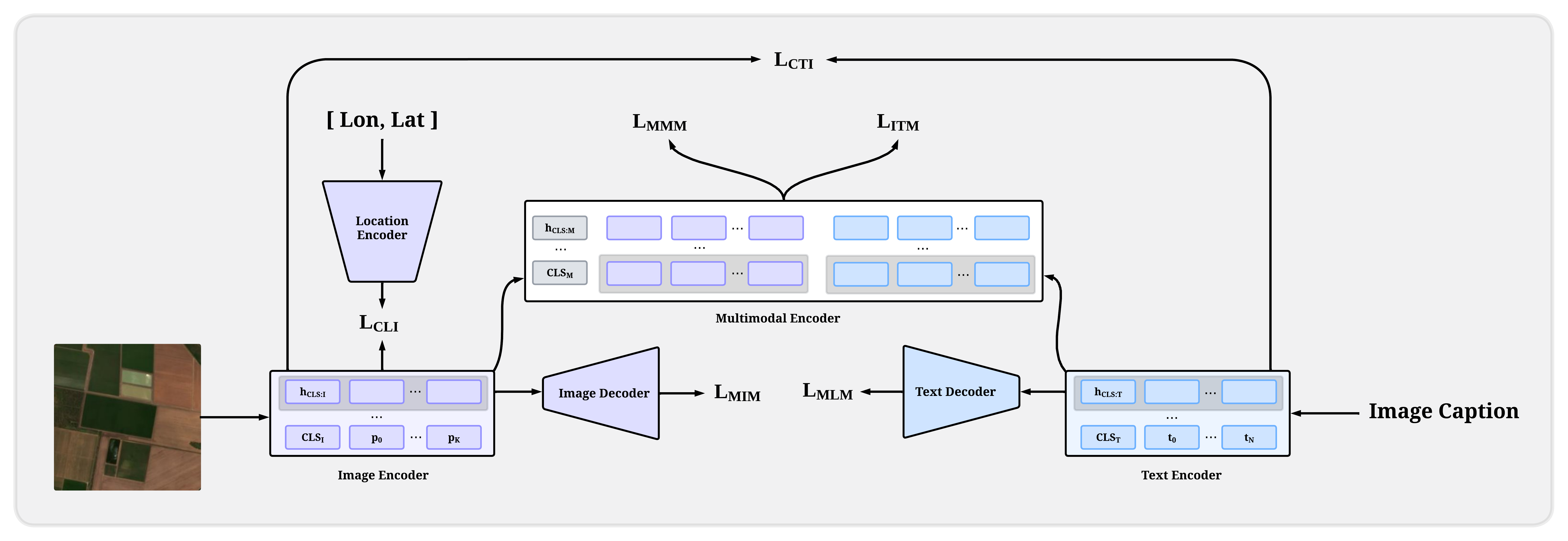}
    \caption{\textbf{The architecture of our proposed FLAVARS vision-language-location pretraining framework.} The components consist of the original FLAVA masked-image modeling, masked-language-modeling, multimodal image-text matching and global image-text contrastive losses. In addition, we combine these with a  geospatial coordinate location-image global contrastive loss which we use to align images, text, and their geospatial coordinates.}
    \label{fig:arch}
\end{figure*}

In the field of remote sensing, imagery is generally dense with objects and visual content which can vary regionally across the globe. This creates a need for vision-language datasets to be highly detailed when describing imagery, and for pretraining to better balance visual task performance while retaining the ability to perform zero-shot classification and image-text retrieval. To better approach multimodal pretraining for remote sensing, we propose \textbf{FLAVARS} -- a multimodal foundation language and visual alignment model for remote sensing. In this work, we study the following questions:

\begin{itemize}
    \item \textbf{Does FLAVA pretraining transfer to the remote sensing domain?} We perform large-scale pretraining of FLAVA on the SkyScript dataset and compare to a SkyScript pretrained CLIP and the original FLAVA pretrained weights.
    \item \textbf{How can we add geospatial-awareness to FLAVA?} We add a global contrastive image-location loss to align image, text, and geospatial location. \textbf{Empirically, we find image-location alignment to be an effective addition to pretraining to improve the downstream performance of the FLAVA encoder representations.}
\end{itemize}

\subsection{Related Work}

\paragraph{Remote Sensing Image Descriptions}
There has recently been an increasing trend in the development of datasets containing remote sensing images and textual descriptions. Manually creating captions for global remote sensing imagery is expensive and time-consuming; therefore, datasets like RS5M~\cite{zhang2023rs5m}, LAION-EO~\cite{czerkawski2023laion}, GeoChat~\cite{kuckreja2024geochat}, and DDFAV~\cite{li2024ddfav} are developed by combining or filtering existing large-scale image-text datasets like LAION-5B~\cite{schuhmann2022laion} for aerial and satellite imagery or generating captions from existing vision task annotations. Alternatively, other works utilize open-source data such as OpenStreetMap (OSM)~\cite{haklay2008openstreetmap} for creating remote sensing image descriptions. OSM is a free resource of rich and diverse geospatial object and image-level annotations of the Earth created manually by a community of global users. SkyScript~\cite{wang2024skyscript} is a dataset of 5 million image-text pairs covering 29k distinct semantic image-level OSM tags.
However, remote sensing images are dense with objects and context, and although these datasets contain image-level text descriptions, other than the RSVG dataset~\cite{zhan2023rsvg} there is a scarcity of research into visual grounding of subregions within each image.

\paragraph{Multimodal Contrastive Learning}
Pretraining with self-supervision is a powerful approach to learning meaningful representations from unlabeled data. In the remote sensing domain, several works have successfully demonstrated success in pretraining with self-supervision across datasets with multiple aligned text or raster modalities. Wang et al.~\cite{wang2022ssl4eo} presented a multimodal and multitemporal Sentinel-1 and 2 dataset providing evidence that contrastive pretraining improves performance across multiple downstream tasks compared to random initialization and ImageNet pretraining. Fuller et al.~\cite{fuller2024croma} and Jain et al.~\cite{jain2022multimodal} apply contrastive pretraining to jointly learn representations from aligned Sentinel-1 radar and Sentinel-2 optical imagery, demonstrating performance improvements over single-modality pretrained models. Liu et al.~\cite{liu2023remoteclip} applied Contrastive Language-Image pretraining (CLIP) to remote sensing by jointly learning from image and text embeddings. CSIP~\cite{corley2022supervising} is a method for joint contrastive learning of remote sensing imagery and height from digital surface models (DSM).

\paragraph{Multimodal Masked Image Modeling}
Mask Autoencoders (MAEs)~\cite{he2022masked} have emerged as an impressive self-supervised learning alternative to contrastive learning methods for the remote sensing domain. SatMAE~\cite{cong2022satmae} presents a pretraining method for temporal and multispectral satellite imagery, supporting the reconstruction of different image modalities. In order to account for multiple scales present in remote sensing, Scale-MAE~\cite{reed2023scale} presents a MAE pretraining method for satellite and aerial imagery at varying ground sampling distances (GSDs) with improved performance when transferring to different resolutions. FG-MAE~\cite{wang2023feature} proposes learning to reconstruct geospatial features such as the Normalized Difference Vegetation Index (NDVI)~\cite{pettorelli2013normalized} from multispectral imagery.

\begin{table*}[t!]
\centering
\resizebox{0.8\textwidth}{!}{%
\begin{tabular}{@{}c|cccccccccccc@{}}
\textbf{Model} &
\rot{\textbf{EuroSAT}} &
\rot{\textbf{RESISC45}} &
\rot{\textbf{PatternNet}} &
\rot{\textbf{RSI-CB-256}} &
\rot{\textbf{AiRound}} &
\rot{\textbf{CvBrCT}} &
\rot{\textbf{MLRSNet}} &
\rot{\textbf{WHU-RS19}} &
\rot{\textbf{OPTIMAL-31}} &
\rot{\textbf{AID}} &
\rot{\textbf{UCMerced}} &
\rot{\textbf{fMoW}} \\
\midrule

SkyCLIP & 94.91 & 91.02 & 98.70 & 99.25 & 75.80 & 76.77 & 94.22 & 97.51 & \textbf{93.01} & 95.75 & \textbf{96.43} & 41.34 \\
FLAVA & 93.17 & 90.03 & 98.22 & 99.19 & 75.46 & 78.47 & 94.56 & \textbf{98.51} & 88.98 & \textbf{95.75} & 94.76 & 39.48 \\
\rowcolor{Gray} FLAVARS (Ours) & \textbf{97.06} & \textbf{92.00} & \textbf{99.23} & \textbf{99.45} & \textbf{78.43} & \textbf{80.22} & \textbf{96.58} & 97.01 & 91.67 & 95.50 & 94.52 & \textbf{42.18} \\
\rowcolor{Gray} + \texttt{LE} & \textbf{97.23} & \textbf{92.81} & \textbf{99.33} & \textbf{99.51} & \textbf{79.08} & \textbf{81.27} & \textbf{96.64} & 97.51 & 91.73 & 95.70 & 95.82 & \textbf{43.37} \\

\bottomrule

\end{tabular}%
}
\caption{
\textbf{K-Nearest Neighbor (\boldmath$k=5$) Classification results} on 12 scene recognition datasets. K-NN classification on image embeddings is an indicator of the frozen representation ability of the vision encoders. Our FLAVARS method pretrained on SkyScript outperforms SkyCLIP on 8/12 datasets. \texttt{LE} represents the addition of Location Encoding objective to the base FLAVA architecture resulting in FLAVARS+\texttt{LE}.
}
\label{tab:knn}
\end{table*}

\section{Methods}
\subsection{SkyScript-Grounded Dataset}
OpenStreetMap (OSM)~\cite{haklay2008openstreetmap} tags are a valuable resource for community-provided annotation of man-made objects and their categorization on the surface of the Earth and have been recently used as a source for generating captions used in remote sensing vision-language datasets. In particular, the SkyScript dataset~\cite{wang2024skyscript} proposed a global and multi-sensor dataset of 5 million image-caption pairs. However, the captions are created using a baseline method of simply combining OSM tags into a short sentence. We build upon this dataset by performing the following improvements to the captions:

\begin{figure}[t!]
    \centering
    \includegraphics[width=0.9\linewidth]{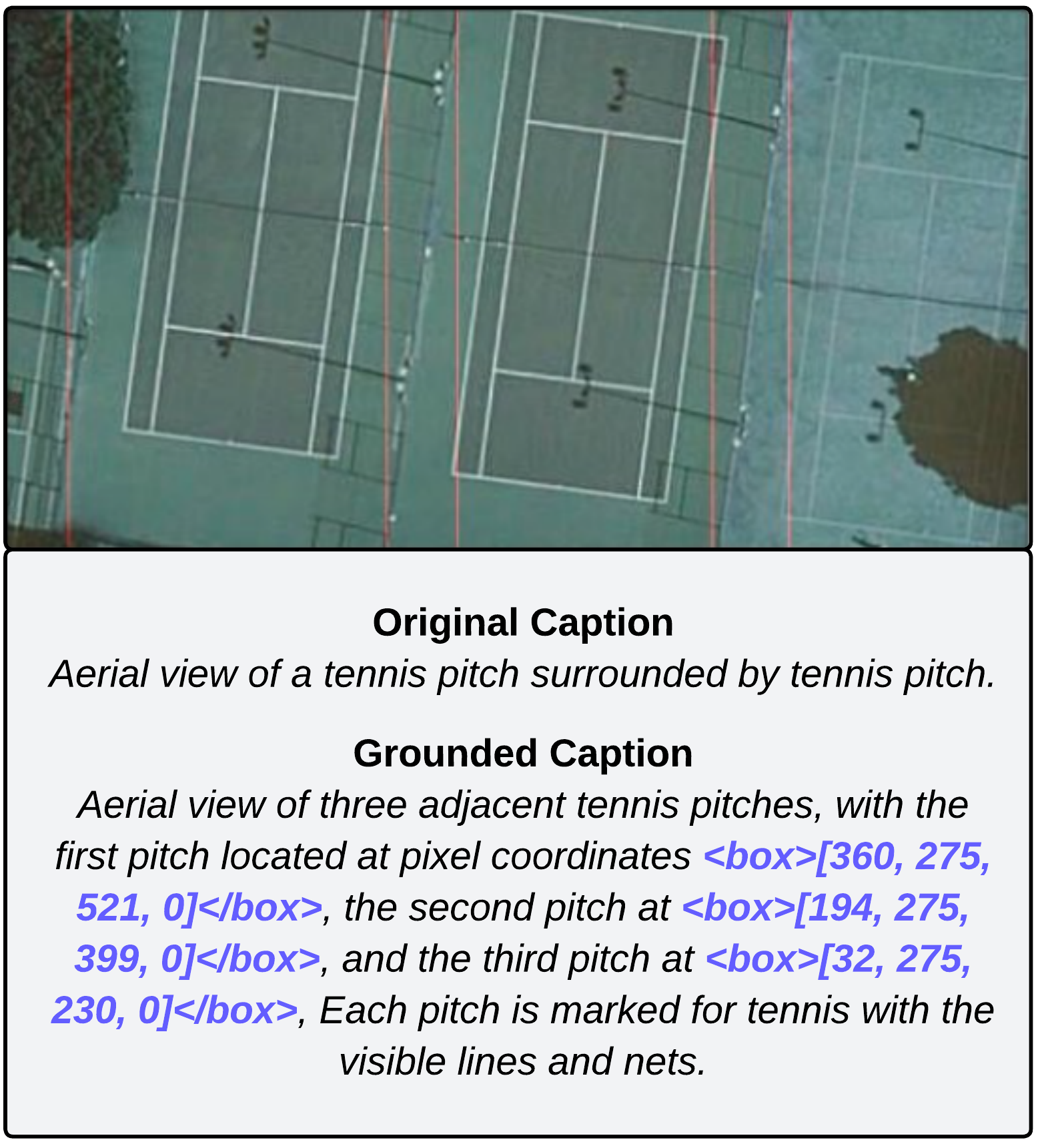}
    \caption{\textbf{A sample from our SkyScript-Grounded dataset.} We improve the original captions in the SkyScript dataset using GPT-4V by prompting with a caption improvement and localization instruction along with the image of interest. The grounded captions contain bounding-box pixel coordinates encompassing the objects in the image with correponding OSM tags. As an example, we plot the resulting boxes on the sample image above.}
    \label{fig:sample}
\end{figure}

\begin{enumerate}
    \item We utilize GPT-4V~\cite{achiam2023gpt} to generate detailed and descriptive paragraph-long captions of imagery by prompting the language model to utilize both the image and original caption for the improved caption generation.
    \item We further incorporate visual grounding into the captions by asking GPT-4V to detect the subjects within the caption and provide structured bounding box coordinates.
\end{enumerate}

We process each image in the SkyScript dataset with the above methods to form the SkyScript-Grounded dataset which contains images, captions, geospatial coordinates, and grounded captions via bounding box annotations. A sample of the dataset is visualized in Figure~\ref{fig:sample}.

\subsection{FLAVARS}
We take inspiration from the FLAVA pretraining framework~\cite{singh2022flava} which incorporates additional masked modeling and vision-language retrieval losses in addition to contrastive learning, with the goal of improving both vision and language downstream performance and alignment.

To improve the image encoder's geospatial representations and global awareness, we modify the FLAVA framework for the remote sensing domain by incorporating a contrastive location-image alignment. Inspired by SatCLIP~\cite{klemmer2023satclip}, we utilize a location encoder initialized with the SatCLIP weights and continually pretrained using the coordinates of the SkyScript dataset. We then constrain the embeddings of the vision encoder to be jointly aligned between image, location, and text. The FLAVARS architecture is provided in Figure~\ref{fig:arch}.

\begin{table*}[t!]
\centering
\resizebox{0.8\textwidth}{!}{%
\begin{tabular}{@{}c|cccccccccccc@{}}

\textbf{Model} &
\rot{\textbf{EuroSAT}} &
\rot{\textbf{RESISC45}} &
\rot{\textbf{PatternNet}} &
\rot{\textbf{RSI-CB-256}} &
\rot{\textbf{AiRound}} &
\rot{\textbf{CvBrCT}} &
\rot{\textbf{MLRSNet}} &
\rot{\textbf{WHU-RS19}} &
\rot{\textbf{OPTIMAL-31}} &
\rot{\textbf{AID}} &
\rot{\textbf{UCMerced}} &
\rot{\textbf{fMoW}} \\
\midrule

SkyCLIP & 37.96 & \textbf{65.67} & \textbf{69.77} & \textbf{44.69} & \textbf{50.70} & 26.81 & \textbf{60.65} & \textbf{88.56} & \textbf{83.06} & \textbf{69.35} & \textbf{73.57} & \textbf{18.49} \\
FLAVA & 23.06 & 42.19 & 44.41 & 24.81 & 44.28 & \textbf{28.57} & 37.44 & 71.64 & 55.38 & 55.45 & 48.81 & 12.93 \\
\rowcolor{Gray} FLAVARS (Ours) & \textbf{43.07} & 38.16 & 48.49 & 23.96 & 48.66 & 20.67 & 38.04 & 63.18 & 56.99 & 49.30 & 49.05 & 11.51 \\
\rowcolor{Gray} + \texttt{LE} & \textbf{41.95} & 37.88 & 48.03 & 23.13 & 48.42 & 20.31 & 37.04 & 62.29 & 53.76 & 49.53 & 46.86 & 11.79 \\

\bottomrule

\end{tabular}%
}
\caption{
\textbf{Zero-Shot Classification results} on 12 scene recognition datasets, assessing the vision-language alignment in latent space. SkyCLIP outperforms our FLAVARS method pretrained on SkyScript on 10/12 datasets, however our method still retains the ability to perform zero-shot classification and text-to-image retrieval while obtaining a better performing visual encoder. Notably, it is expected that CLIP pretraining should fare well here as alignment is the only objective during pretraining. FLAVARS, on the other hand, has multiple loss functions seeking to improve not only alignment but text and vision encoder individual performance. \texttt{LE} represents the addition of Location Encoding objective.
}
\label{tab:zero-shot}
\end{table*}

\section{Experiments}
\label{sec:experiments}

\subsection{Pretraining}
We proceed by pretraining FLAVA and our FLAVARS method on the OpenAI CLIP\cite{radford2021learning} top-30\% scoring subset of the SkyScript dataset~\cite{wang2024skyscript} to fairly compare to the SkyCLIP~\cite{wang2024skyscript} weights trained on the same subset. We note that while we introduce the SkyScript-Grounded dataset, we leave the investigation of its performance for future work. 

\subsection{Benchmark Datasets}
We perform a thorough analysis of the performance of SkyCLIP, FLAVA, and our proposed FLAVARS methods using 12 remote sensing scene recognition datasets and the SpaceNet1 semantic segmentation dataset~\cite{van2018spacenet}. Furthermore, we find that several prior works appear to use different random splits for each dataset without publishing the seed used or filenames in each set. We correct this by publishing our exact splits used for training, validation, and test sets, such that the community can reproduce our work\footnote{\href{https://huggingface.co/datasets/isaaccorley/FLAVARS}{huggingface.co/datasets/isaaccorley/FLAVARS}}.

\subsection{KNN Image Classification}
We perform experiments by embedding the images in all image classification datasets utilizing the frozen vision encoder of each method. Results are provided in Table~\ref{tab:knn}. Our results indicate that the FLAVARS pretraining significantly improves over CLIP-based pretraining by enabling the vision encoder to produce representations which can be easily classified in Euclidean space. Furthermore, we notice that adding location encoding alignment provides an additional increase in performance compared to the base FLAVA architecture.

\subsection{Zero-Shot Image Classification}
We perform experiments by embedding the images and textual class labels in all image classification datasets utilizing the vision and text encoders for each method. We then classify by selecting the class label with the highest cosine similarity to the image embedding. Results are provided in Table~\ref{tab:zero-shot}. Our results indicate that CLIP-based pretraining significantly improves over FLAVARS-based pretraining by producing better vision-language alignment. However, this is expected, as this objective is the primary loss used in CLIP pretraining, whereas FLAVARS has several loss functions that seek to balance and improve all representations.

\subsection{Semantic Segmentation}
To further assess intermediate image layer representations, we combine each pretrained vision encoder with a UperNet decoder~\cite{xiao2018unified} to form a semantic segmentation architecture. We train on the SpaceNet1 building binary segmentation dataset. Results are provided in Table~\ref{tab:spacenet}. We further compare to other vision-only baselines not pretrained with multiple modalities -- ImageNet and MAE. Our results indicate that our multimodal FLAVARS pretraining significantly improves dense vision performance over CLIP pretraining and performs well compared to other baselines.

\begin{table}[t!]
\centering
\resizebox{0.43\textwidth}{!}{%
\begin{tabular}{@{}ccccc@{}}

\toprule
\textbf{Method} & \textbf{Model} & \textbf{Backbone} & \textbf{Pretrain Dataset} & \textbf{mIoU} \\
\toprule

Supervised & U-Net & ResNet-50 & ImageNet & 76.5 \\
MAE & UperNet & ViT-B16 & SkyScript & 76.7 \\
MAE & UperNet & ViT-B16 & ImageNet & 77.8 \\

\midrule\midrule
SkyCLIP & UperNet & ViT-L14 & SkyScript & 72.0 \\
\rowcolor{Gray} FLAVARS & UperNet & ViT B16 & SkyScript & \textbf{77.9} \\
\rowcolor{Gray} + \texttt{LE} & UperNet & ViT B16 & SkyScript & \textbf{78.1} \\

\bottomrule
\end{tabular}%
}
\caption{\textbf{SpaceNet v1 Semantic Segmentation results.} Our FLAVARS method pretrained on SkyScript outperforms SkyCLIP on dense prediction tasks. We further compare to vision only pretraining methods MAE and ImageNet pretrained weights. \texttt{LE} represents the addition of the Location Encoding objective.}
\vspace{-5mm}
\label{tab:spacenet}
\end{table}

\section{Discussion \& Conclusion}
\label{sec:discussion-conclusion}
In this work, we proposed FLAVARS, a new multimodal foundation model for remote sensing. Building on existing models such as CLIP and FLAVA, FLAVARS demonstrates improvement in representation ability by combining masked-image \& masked-language modeling with contrastive learning for remote sensing.

In Table~\ref{tab:knn} our results indicate FLAVARS is able to outperform vision-only pretrained methods like MAE~\cite{he2022masked} while retaining vision-language and vision-location alignment for zero-shot classification and retrieval. While in Table~\ref{tab:zero-shot}, CLIP outperforms FLAVARS, this is unsurprising due to the zero-shot image-text classification task being precisely the CLIP objective. FLAVARS, however, notably outperforms FLAVA in all cases for KNN classification of image embeddings and in most cases for zero-shot classification. We further find that spatial understanding can be improved by aligning image and text representations with geographic coordinates shown by the added location encoding contrastive objective. Lastly, we provide evidence there is still a noticeable trade-off in performance of vision-language pretraining across dense visual tasks, like segmentation, and multimodal alignment capabilities, e.g. zero-shot classification. Future research is necessary to investigate this and how to systematically improve upon this trade-off.

\section*{Acknowledgements}
We thank the Microsoft AI for Good Research Lab for providing the compute and resources required to perform the experiments in this paper.

{\small
\bibliographystyle{ieee_fullname}
\bibliography{egbib}
}

\end{document}